\documentclass[letterpaper]{article} % DO NOT CHANGE THIS
\usepackage{aaai24}  % DO NOT CHANGE THIS
\usepackage{times}  % DO NOT CHANGE THIS
\usepackage{helvet}  % DO NOT CHANGE THIS
\usepackage{courier}  % DO NOT CHANGE THIS
\usepackage[hyphens]{url}  % DO NOT CHANGE THIS
\usepackage{graphicx} % DO NOT CHANGE THIS
\usepackage{natbib}  % DO NOT CHANGE THIS AND DO NOT ADD ANY OPTIONS TO IT
\usepackage{caption} % DO NOT CHANGE THIS AND DO NOT ADD ANY OPTIONS TO IT
\usepackage{algorithm}
\usepackage{algorithmic}

\usepackage{newfloat}
\usepackage{listings}
\DeclareCaptionStyle{ruled}{labelfont=normalfont,labelsep=colon,strut=off} % DO NOT CHANGE THIS
\floatstyle{ruled}
\newfloat{listing}{tb}{lst}{}
\floatname{listing}{Listing}
\nocopyright
\title{Bidirectional Contrastive Split Learning for Visual Question Answering}
\author{
    Yuwei Sun \textsuperscript{\rm 1,2},
    Hideya Ochiai \textsuperscript{\rm 1}
}
\affiliations{
    \textsuperscript{\rm 1}University of Tokyo\\
    \textsuperscript{\rm 2}RIKEN AIP\\
    ywsun@g.ecc.u-tokyo.ac.jp
}

\usepackage{subcaption}
\usepackage{comment}
\usepackage{tikz}
\def\checkmark{\tikz\fill[scale=0.4](0,.35) -- (.25,0) -- (1,.7) -- (.25,.15) -- cycle;} 
\usepackage{amsmath,amsfonts,amssymb}
\usepackage{dsfont}
\usepackage{multirow}

\usepackage{bibentry}
\begin{document}

\maketitle

\begin{abstract}
Visual Question Answering (VQA) based on multi-modal data facilitates real-life applications such as home robots and medical diagnoses. One significant challenge is to devise a robust decentralized learning framework for various client models where centralized data collection is refrained due to confidentiality concerns. This work aims to tackle privacy-preserving VQA by decoupling a multi-modal model into representation modules and a contrastive module and leveraging inter-module gradients sharing and inter-client weight sharing. To this end, we propose Bidirectional Contrastive Split Learning (BiCSL) to train a global multi-modal model on the entire data distribution of decentralized clients. We employ the contrastive loss that enables a more efficient self-supervised learning of decentralized modules. Comprehensive experiments are conducted on the VQA-v2 dataset based on five SOTA VQA models, demonstrating the effectiveness of the proposed method. Furthermore, we inspect BiCSL's robustness against a dual-key backdoor attack on VQA. Consequently, BiCSL shows much better robustness to the multi-modal adversarial attack compared to the centralized learning method, which provides a promising approach to decentralized multi-modal learning.
\end{abstract}

\section{Introduction}

The deployment of multi-modal models in safety-critical applications, such as personal robots and healthcare, requires addressing robust architecture design. The collected vast amount of user data causes critical privacy concern. Unfortunately, few studies have focused on enhancing privacy for multi-modal models. For instance, Visual Question Answering (VQA) requires a large amount of data in both texts and images that indicate a wide range of personal interests. Decentralized machine learning, such as federated learning (FL), is one of the approaches to privacy-preserving VQA through the collaborative learning of different local models via weight sharing. Conventional FL methods \cite{fedavg} for VQA tasks have two main drawbacks: 1) models trained on separate client data are aggregated with model parameter sharing. However, sharing a complete model might lead to adversarial attacks \cite{patch4}; 2) training a large model on resource-constrained client devices could be inefficient and impractical.

\begin{figure*}
    \centering
    \includegraphics[width=0.85\textwidth]{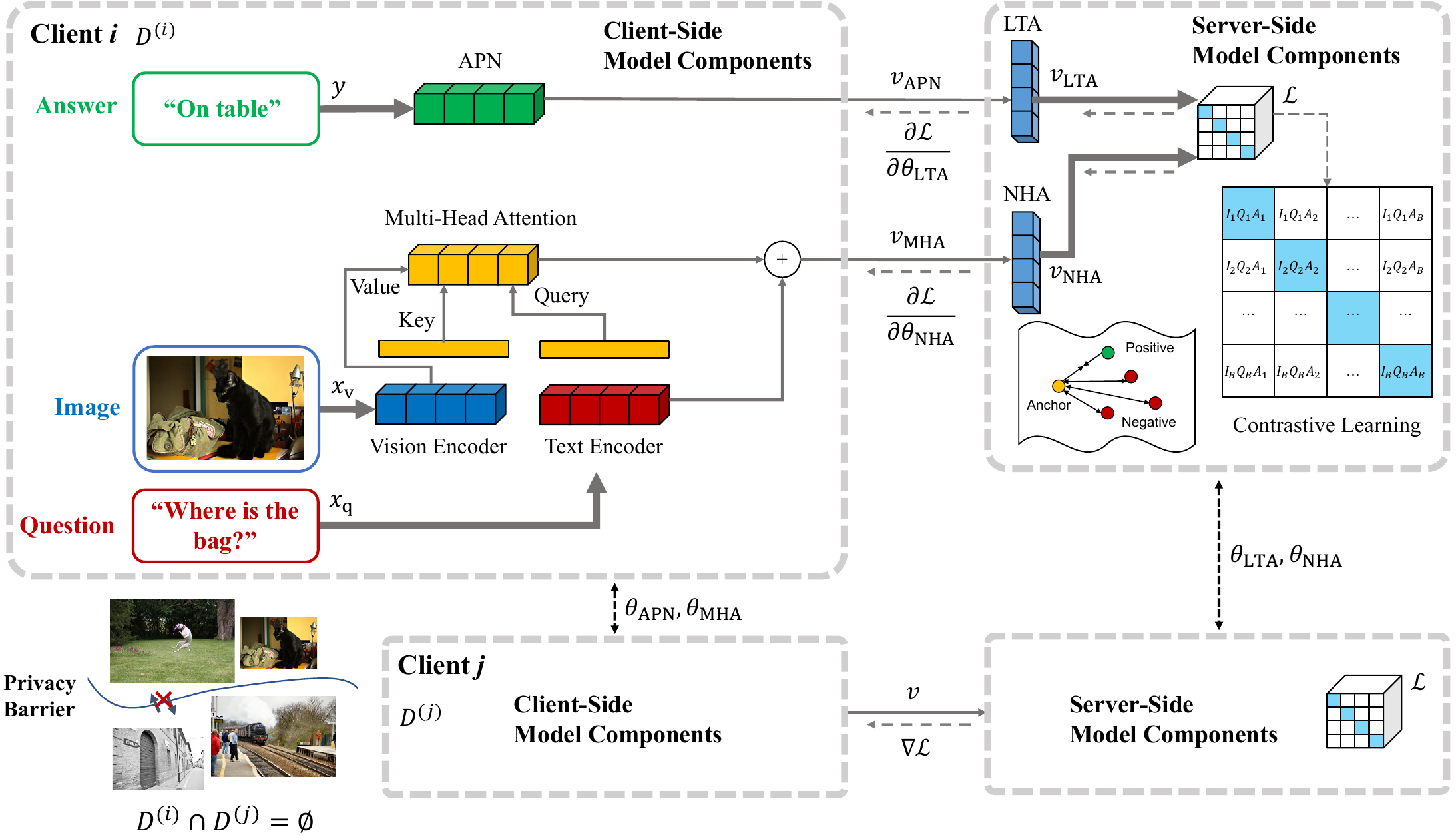}
    \caption{BiCSL for decentralized visual question answering consists of three main components: cross-modal representation learning (multi-head attention), an answer projection network (APN) for semantic understanding of answers, and two adapter networks (LTA and NHA) for contrastive learning of different model component outputs. BiCSL learns refined representations from different clients via inter-client weight sharing, while ensuring privacy protection via inter-module gradient sharing.}
    \label{fig:scheme}
\end{figure*}

We aim to overcome the aforementioned challenges by proposing the Bidirectional Contrastive Split Learning (BiCSL) method. Different from FL which trains the entire model on a local device, BiCSL decouples a large-scale model into client components and cloud components. This avoids the potential misuse of the exposed model architecture by attackers, such as mounting a backdoor attack using the revealed architecture. BiCSL learns refined cross-modal representations from different clients via inter-module gradient sharing and inter-client weight sharing, without revealing either the user data or the model architecture (Figure \ref{fig:scheme}). This is enabled by a self-supervised learning method to correlate various decentralized modules.

The main contributions of this work are as follows:

1) We propose a novel self-supervised split learning method for VQA, called Bidirectional Contrastive Split Learning (BiCSL). BiCSL trains a global model over the entire client data distribution without disclosing either training data or model architecture. This is the first study of self-supervised decentralized VQA. 

2) This study demonstrates BiCSL's ability to tackle self-supervised learning of decentralized multi-modal data. BiCSL devises a contrastive learning method to align module activations encouraging similarity between relevant outputs while discouraging similarity between irrelevant ones.
 
3) An in-depth evaluation with a wide range of metrics including robustness to adversarial attack is conducted. The results show that our method could achieve competitive performance compared to a centralized method, while ensuring privacy protection and maintaining great performance even under adversarial attacks.

\section{Related Work}

\subsection{Decentralized Machine Learning}

Decentralized Machine Learning (DML) \cite{survey} encompasses methods such as Federated Learning (FL) \cite{fedavg, chaoyang, fedmd}, Split Learning (SL) \cite{split}, and Swarm Learning \cite{swarm}. These methods address privacy concerns by enabling collaborative learning without the need for centralized data storage. Although DML has been widely investigated for single-modality tasks, its application to multi-modal models is still limited. For example, aimNet \cite{fedvqa} is a FL-based VQA framework, which utilizes fine-grained representations from various clients to enhance downstream tasks. Unfortunately, aimNet is a supervised method relying on annotated answer labels. Additionally, sharing client models during training increases its vulnerability to adversarial attacks.

\begin{table}[!t]
    \centering
    \renewcommand{\arraystretch}{1.2}
    \resizebox{\linewidth}{!}{%
    \begin{tabular}{lcccc}
    \hline
    Methods & Shared Data & Shared Model  & Learning Framework & Loss Function\\
    \hline
    MMNas  &  $\checkmark$ & $\checkmark$ & Single fusion & Cross entropy\\
    QICE & $\checkmark$ & $\checkmark$ & Single fusion & Contrastive loss\\
    aimNet & $\times$ & $\checkmark$ & Federated Learning & Cross entropy\\
    BiCSL (Ours) & $\times$ & $\times$ & Split Leaning & Contrastive loss\\
    \hline
    \end{tabular}%
    }
    \caption{Comparison of VQA methods: BiCSL does not require sharing training data or models. Different from previous work on decentralized VQA of the aimNet, BiCSL is a self-supervised method without the need for training labels.}
    \label{table:related}
\end{table}

\subsection{Visual Question Answering}
Multi-modal machine learning (MMML) \cite{mmv,clip,avlnet,dalle,dalle2} has been intensively studied to understand across different modalities of information. A specific task within MMML is Visual Question Answering (VQA) \cite{butd,mmnas}, which involves answering natural language questions based on the contents of a presented image. Nevertheless, the vast majority of VQA studies so far rely on modality fusion methods where VQA is considered a centralized multi-class classification task. Moreover, previous studies usually do not consider the privacy concerns associated with centralized large-scale model training (Table \ref{table:related}).

Contrastive learning is an alternative to the supervised method, which computes a cosine similarity matrix among all possible candidates of images and texts within a batch. For instance, Question-Image Correlation Estimation (QICE) \cite{zhu20} aims to train on relevant image and question pairs in VQA datasets to alleviate the language prior problem \cite{prior2}. Nevertheless, QICE does not provide any guarantees regarding data or model privacy. In contrast, we found that contrastive learning could be used as a natural fit for privacy-preserving VQA by consolidating with decentralized learning techniques.

\section{Methods}

In this section, we delve into a comprehensive exploration of the proposed method's technical underpinnings. These include the incorporation of split learning for decentralized VQA, an answer projection network for enhanced understanding of semantic notions, a contrastive learning architecture for effective training on unlabeled client data, and inter-client weight sharing for local update aggregation.

\subsection{Attention-Based VQA}
Visual Question Answering (VQA) is a task that involves answering natural language questions based on the visual content of a given image. Typically, the VQA problem is approached as a supervised learning task with a predetermined list of $C$ potential answer options. Let $f_{\text{MHA}}$ be the VQA model that takes as the input the pair of an image $x_v\in \mathbb{R}^V$ and a question $x_q\in \mathbb{R}^Q$ and outputs an answer $\hat{y} \in \{y_1,y_2,...,y_C\}$ where $y_c \in \mathbb{R}^A$. A VQA model aims to predict the correct answer $y$ given the input pair $(x_v,x_q) \in D$ where $D$ is the dataset. $\hat{y}=\underset{y}{\mbox{arg max}}\,p(y|x_v,x_q;f_{\text{MHA}})$ where $p(\cdot|\cdot)$ is the conditional probability. 

Moreover, we study a diverse set of VQA models that are based on the attention mechanism \cite{attention2}. Cross-attention in VQA models enables improved refined representation learning from multi-modal data. At its simplest form, each head of a multi-head attention (MHA) maps a query and a set of key-value pairs to an output. Let $W^{\text{t}} \in \mathbb{R}^{Q\times M}$ be an encoder to process the text input $x_q$ (such as LSTM \cite{lstm} and Transformer \cite{attention2}), and $W^{\text{v}} \in \mathbb{R}^{V\times P}$ be an encoder to process the vision input $x_v$ (such as CNN \cite{cnn} and MLP \cite{mlp}). The linearly projected output of the text encoder is used as a query $Q \in \mathbb{R}^d \leftarrow W^{Q_i}W^{\text{t}}x_q)$, which is compared with that of the vision encoder which serves as the key $K \in \mathbb{R}^d \leftarrow W^{K_i}W^{\text{v}}x_v$. Here, $W^Q \in \mathbb{R}^{M\times d}$ and $W^K \in \mathbb{R}^{P\times d}$ are linear transformations for the query and key. Then, the weighted sum of values $V \in \mathbb{R}^{P} \leftarrow W^{\text{v}}x_v$ could be formulated as follows
$$\mbox{h}^i(x_v,x_q)=\mbox{softmax}(\frac{W^{Q_i}W^{\text{t}}x_q(W^{K_i}W^{\text{v}}x_v)^T}{\sqrt{d}})\,W^{\text{v}}x_v,$$

\begin{equation}\mbox{Multi-head}(x_v,x_q)=\mbox{Concat}(\mbox{h}^1,\dots,\mbox{h}^H)\,W^O,
\end{equation}
where $W^O$ is a linear transformation for outputs, and $H$ is the number of attention heads.

\begin{figure}[!t]
    \raggedright
    \includegraphics[width=\linewidth]{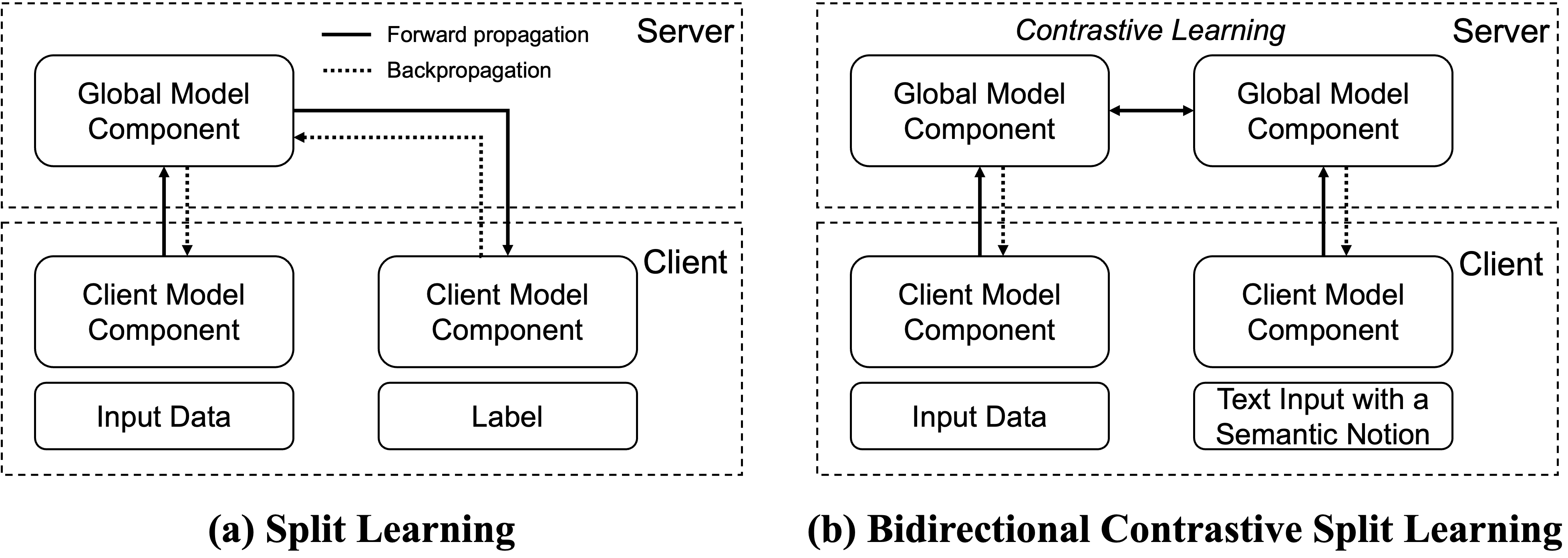}
    \caption{Conventional Split Learning vs. BiCSL: (a) Split Learning utilizes numeric one-hot vectors of answer labels for model training, based on a unidirectional process that requires sequential processing of components resulting in longer waiting time. (b) BiCSL employs lexical semantic notions of answer texts and a bidirectional process that enables concurrent processing of model components.}
    \label{fig:compare}
\end{figure}

\subsection{Decentralized VQA}
To devise a decentralized method, training numerous VQA models on client devices is inefficient and impractical due to resource constraints on local devices. Intuitively, we could divide a complete model into client and server components. Then, by leveraging inter-module gradient sharing, the parameters of each component could be efficiently updated and synchronized. To this end, we consider dividing a VQA model into three components, i.e., a global component $f_{g}$ and two client components $\{f_{c,1},f_{c,2}\}$ (Figure \ref{fig:compare}.a). Then, we assume that $K$ client models are trained on their local datasets $D^{(k)}$, which consist of $N^{(k)}$ samples, represented as $\{(x_{v,j}, x_{q,j}, y_j)\}_{j=1}^{N^{(k)}}$. Here, $\cup_{k=1}^K D^{(k)} = D$, $D^{(i)} \cap D^{(j)} = \emptyset,,\forall i\neq j$, and $\sum^{K}_{k=1} N^{(k)} = N$, where $N$ is the sample size. Furthermore, we make the assumption that the client models share the same architecture, and the division of the models are consistent across all clients. Training data sharing among clients is not possible due to confidentiality.

Then, the decentralized VQA method proceeds by iterating the following steps: (1) each client $k$ computes the output of the component $f_{c,1}$ with $D^{(k)}$ and sends the output to the server, (2) the server forward-propagates the input with the global component $f_{g}$ and sends back the output, (3) the probability distribution and the loss are computed by $f_{c,2}$ using ground-truths $\{y_j\}_{j=1}^{N^{(k)}}$, (4) the gradients $(\delta_k \theta_{c,1}, \delta_k \theta_{c,2}, \delta_k \theta_{g})$ for each component of client $k$ are then computed via an inverse path $f_{c,2}\rightarrow f_{g} \rightarrow f_{c,1}$, (5) after all clients complete local training, their update gradients are averaging aggregated for inter-client weight sharing, $\delta \theta_{c,1} = \frac{1}{K}\underset{k \in K} {\sum}\delta_k \theta_{c,1} $, $\delta \theta_{c,2} = \frac{1}{K}\underset{k \in K} {\sum}\delta_k \theta_{c,2} $, $\delta \theta_{g} = \frac{1}{K}\underset{k \in K} {\sum}\delta_k \theta_{g}$, and (6) the aggregated updates are distributed to clients for the update of their local components. We repeat the process above until a global training goal is achieved. This architecture enables clients to train individual models without sharing local data or models, while harnessing the acquired knowledge from other clients through activation and gradient sharing.

\subsection{Bidirectional Contrastive Split Learning}

Though the aforementioned supervised decentralization of VQA enhances privacy of local model training, there exist two main drawbacks. First, the semantic understanding of answers is often misaligned with the inputs due to the image and question pairs are labeled with numeric ids of answer texts. Second, the computational time could be substantial due to the interactive activation and gradient sharing among components. To this end, we propose a self-supervised decentralization method for VQA, called Bidirectional Contrastive Split Learning (BiCSL). BiCSL leverages contrastive learning-based component alignment to enhance the correlation between visual and language contents and improve efficiency of activation and gradient sharing.

\subsubsection{Answer Projection and Adapter Networks}
An Answer Projection Network (APN) $f_{\text{APN}}$ (Figure \ref{fig:arch}.c) aims to project a lexical answer $y$ into a feature vector $v_{\text{APN}}\in \mathbb{R}^S$. APN comprises two main components: a preprocessing process and the word embedding of GloVe \cite{pennington2014glove} to transform the question text into a fixed-size vector representation. The resultant vector is subsequently fed through a linear projection layer.

Moreover, two adapter networks (Figure \ref{fig:arch}.a, \ref{fig:arch}.b) are employed to project the outputs of client components into a shared dimension, where a Nonlinear Head Adapter (NHA) is applied to tackle more complex representations while a Linear Tail Adapter (LTA) is used to process simpler ones \cite{mmv}. In particular, to tailor a VQA model for contrastive learning, we replace its output layer with the NHA network $f_{\text{NHA}}$. The NHA projects the learned cross-modal representations into $v_{\text{NHA}} \in \mathbb{R}^S$. We use the LTA network $f_{\text{LTA}}$ to project the learned representations from the APN $v_{\text{APN}}$ into $v_{\text{LTA}} \in \mathbb{R}^S$. Note that $v_{\text{LTA}}$ and $v_{\text{NHA}}$ have the same dimension of $S$.

\subsubsection{Contrastive Learning of Model Components}

\begin{figure}[!t]
\centering
    \includegraphics[width=\linewidth]{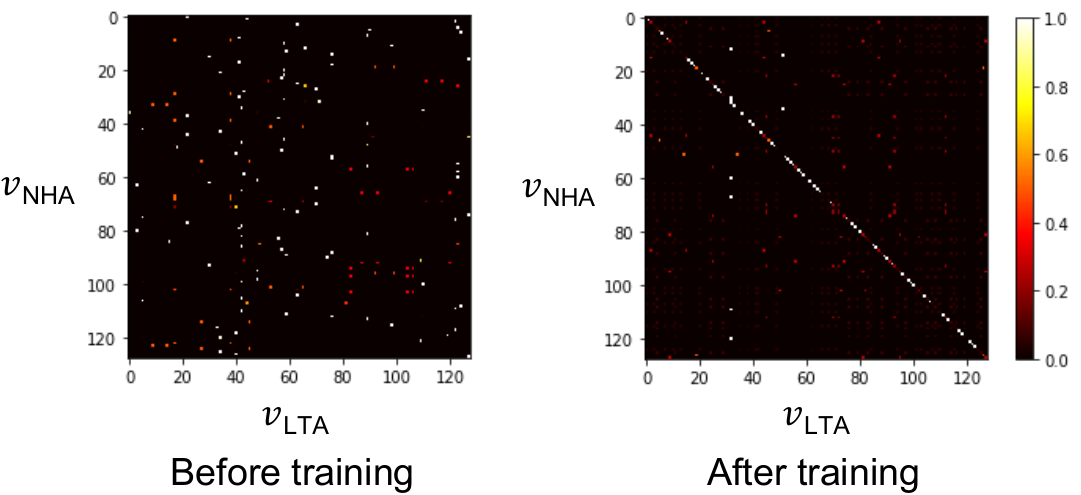}
    \caption{Measured dot product similarity between any two representations in one batch before and after training with BiCSL. The similarity scores are optimized such that the representations of positive pairs have a higher score while that of the negative pairs have a lower score.}
    \label{fig:heat}
\end{figure}

We employ the Information Noise Contrastive Estimation (InfoNCE) loss \cite{infonce} to disentangle similar (positive) and dissimilar (negative) pairs of data points (Figure \ref{fig:heat}). Model component activations are aligned for the positive pairs while being discouraged for the negative pairs. Notably, we use the NHA and LTA outputs for the same inputs as the positive pairs, i.e., $\{(v_{\text{NHA},j},v_{\text{LTA},j})\}_{j=1}^B$ where $B$ is the batch size. On the contrast, given the NHA output $v_{\text{NHA},i}$, the irrelevant LTA outputs $\{v_{\text{LTA},j}|j\neq i\}_{j=1}^B$ within one batch are employed as the negative pairs. Consequently, we devise the loss $\mathcal{L}$ for the contrastive learning of model components as follows 

\begin{equation}
        \mathcal{L}= -\sum_{i=1}^{B}\log\frac{\exp(v_{\text{NHA},i} \cdot v_{\text{LTA},i} / \tau)}{\sum_{j=1}^{B} {1}_{[j \neq i]} \exp(v_{\text{NHA},i} \cdot v_{\text{LTA},j} / \tau)},
\end{equation}
where $\tau$ is the temperature parameter to ensure the output is appropriately scaled to the data distribution, and ${1}_{[j \neq i]}$ is an indicator function: 1 if $j \neq i$, 0 otherwise. 

Additionally, the proposed framework enables a parallel processing of model components improving the efficiency of decentralized VQA. In the decentralized VQA based on split learning, the activation and gradient sharing between the client and the server is unidirectional (Figure \ref{fig:compare}.a). Each component needs to process the input data in subsequent which could largely increase the waiting time during training. In contrast, in our architecture, all layer activations are sent from clients to the server while all gradients are sent from the server to clients. As a result, clients could utilize their local components concurrently while computing activations or gradients, without waiting for the computation of the previous component (Figure \ref{fig:compare}.b).

\subsubsection{Local Update Aggregation}
Every epoch $t$, aggregating model updates $\theta_{t+1}^{(k)}-\theta_t^{(k)}$ from different clients $k\in \{1,2,\dots,K\}$ enhances the generality of the aggregated global model. Due to sending client updates to the server for aggregation could expose the model architecture, we employ a dual-server parameter aggregation approach that leverages a second auxiliary server for the aggregation of client updates (APN and MHA). The aggregation of the server updates (NHA and LTA) is performed on the main server. We use an averaging aggregation method formulated as follows

\begin{equation}
 \delta \theta_{t} = \frac{1}{K}\underset{k\in \{1,2,\dots,K\}}{\sum}(\theta_{t+1}^{(k)}-\theta_t^{(k)}),  
\end{equation}
where $\theta$ is the parameters of a model component from $\{\theta_{\text{APN}},$ 
$ \theta_{\text{MHA}}, \theta_{\text{NHA}}, \theta_{\text{LTA}}\}$. 

The proposed BiCSL method is demonstrated in Algorithm \ref{BiCSL}. 

\begin{algorithm}[!t]
\caption{BiCSL}
\label{BiCSL}
\begin{algorithmic}[1]
\STATE $T$: number of rounds
\STATE $E$: number of local epochs
\STATE $\eta$: learning rate

\FOR {each round $t = 1, 2, \dots, T$}
    \FOR {each client $k\in \{1,2,\dots,K\}$ in parallel}
       \FOR{$\theta \in \{\theta_{\text{APN}},\theta_{\text{MHA}}, \theta_{\text{NHA}}, \theta_{\text{LTA}}\}$} 
       \STATE $\theta_{t}^{(k)}\leftarrow \theta_{t}$
       \ENDFOR
       \FOR{each local epoch $e = 1, 2,\dots, E$}
            \STATE $v_{\text{MHA},t,e}^{(k)}$ = $f_{\text{MHA}}(\theta_{\text{MHA},t,e}^{(k)},(x_v^{(k)}, X_q^{(k)}))$
            \STATE $v_{\text{APN},t,e}^{(k)}$ = $f_{\text{APN}}(\theta_{\text{APN},t,e}^{(k)},Y^{(k)})$
            \STATE $\delta_{\text{NHA},e}^{(k)},\delta_{\text{LTA},e}^{(k)}$ = Server($v_{\text{MHA},t,e}^{(k)}$,$v_{\text{APN},t,e}^{(k)}$)
            \STATE $\theta_{\text{MHA},t,e+1}^{(k)} \leftarrow \theta_{\text{MHA},t,e}^{(k)}-\eta\cdot \frac{\partial \delta_{\text{NHA},e}^{(k)}}{\partial \theta_{\text{MHA},t,e}^{(k)}}$
            \STATE $\theta_{\text{APN},t,e+1}^{(k)} \leftarrow \theta_{\text{APN},t,e}^{(k)}-\eta\cdot \frac{\partial \delta_{\text{LTA},e}^{(k)}}{\partial \theta_{\text{APN},t,e}^{(k)}}$\
        \ENDFOR
   \ENDFOR
   \FOR{$\theta \in \{\theta_{\text{APN}},\theta_{\text{MHA}}, \theta_{\text{NHA}}, \theta_{\text{LTA}}\}$} 
   \STATE $\theta_{t+1} = \frac{1}{K}\sum_{k\in K}\theta_{t,E+1}^{(k)}$
   \ENDFOR
\ENDFOR
\STATE
\STATE \textbf{function} Server($v_{\text{MHA},t,e}^{(k)}$,$v_{\text{APN},t,e}^{(k)}$)
\STATE $v_{\text{NHA},t,e}^{(k)} \leftarrow f_{\text{NHA}}(v_{\text{MHA},t,e}^{(k)})$  
\STATE $v_{\text{LTA},t,e}^{(k)} \leftarrow f_{\text{LTA}}(v_{\text{APN},t,e}^{(k)})$  
\STATE $\mathcal{L}= -\sum_{i=1}^{B}\log\frac{\exp(v_{\text{NHA},i} \cdot v_{\text{LTA},i} / \tau)}{\sum_{j=1}^{B} {1}_{[j \neq i]} \exp(v_{\text{NHA},i} \cdot v_{\text{LTA},j} / \tau)}$
\STATE $\delta_{\text{NHA},e}^{(k)}=\frac{\partial\mathcal{L}}{\partial \theta_{\text{NHA},t,e}^{(k)}}$\\
\STATE $\delta_{\text{LTA},e}^{(k)}=\frac{\partial\mathcal{L}}{\partial \theta_{\text{LTA},t,e}^{(k)}}$\\
\STATE $\theta_{\text{NHA},t,e+1}^{(k)}\leftarrow \theta_{\text{NHA},t,e}^{(k)}-\eta\cdot \delta_{\text{NHA},e}^{(k)}$
\STATE $\theta_{\text{LTA},t,e+1}^{(k)}\leftarrow \theta_{\text{LTA},t,e}^{(k)}-\eta\cdot \delta_{\text{LTA},e}^{(k)}$
\RETURN $\delta_{\text{NHA},e}^{(k)},\delta_{\text{LTA},e}^{(k)}$ to client $k$
\end{algorithmic}
\end{algorithm}

\section{Experiments}

In this section, we provide a detailed description of the datasets, model architectures, and metrics used in the experiments. An extensive empirical evaluation is performed based on five SOTA VQA models. Furthermore, we investigate BiCSL's robustness to a sophisticated dual-key backdoor attack on VQA models, comparing its performance against different methods. The results demonstrate that BiCSL achieves competitive performance to the centralized method and remains effective even under the mounted attack.

\subsubsection{Dataset} 
%Our method is evaluated on the benchmark dataset VQA-v2 \cite{vqabase} where images are from the COCO dataset \cite{coco}, including colorful 82,783 training images and 40,504 validation images with a size of 640×480. VQA-v2 contains 443,757 questions for training and 214,354 questions for validation. We report the results on its validation split in this work using Eq. \ref{eq:acc}.

Our method was evaluated on the benchmark dataset VQA-v2 \cite{vqabase} with varying partitioning configurations for decentralized VQA. 
%CLEVR is a synthetic dataset with 3D-rendered objects including 70k images and 700k questions for training and 15k images and 150k questions for validation. 
VQA-v2 covers 82.8k images and 443.8k questions for training and 40.5k images and 214.4k questions for validation. The images are from the COCO dataset \cite{coco} with a size of 640×480. Depending on the client number, we separated the training dataset into several non-overlapping subsets as client datasets. Moreover, we used the entire validation dataset to evaluate the performance of the aggregated global model.

\subsubsection{VQA Models} \label{sec:models} The following VQA models were studied: (1) Multi-modal Factorized Bilinear (MFB) \cite{mfb} combines multi-modal features using an end-to-end network architecture to jointly learn the image and question attention, (2) Bottom-Up and Top-Down attention mechanism (BUTD) \cite{butd} enables attention to be calculated at the level of objects and other salient image regions. The bottom-up mechanism based on Faster R-CNN proposes image regions, while the top-down mechanism determines feature weightings, (3) Bilinear Attention Networks (BAN) \cite{ban} considers bilinear interactions among two groups of input channels and extracts the joint representations for each pair of channels, (4) Multi-modal neural architecture search (MMNas) \cite{mmnas} uses a gradient-based algorithm to learn the optimal architecture, and (5) Modular Co-Attention Network (MCAN) \cite{mcan} consists of Modular Co-Attention layers cascaded in depth where each layer models both the self-attention and the guided-attention of the input. 

We evaluated the model performance with three different seeds and reported the mean and standard deviation. The VQA models were implemented using PyTorch with their default hyperparameters. The experiments were conducted on four A100 GPUs with 40GB memory. The code would be made publicly available.

\subsubsection{Architecture and Hyperparameters}
\label{structure}

\begin{figure}[!t]
    \centering
    \includegraphics[width=\linewidth]{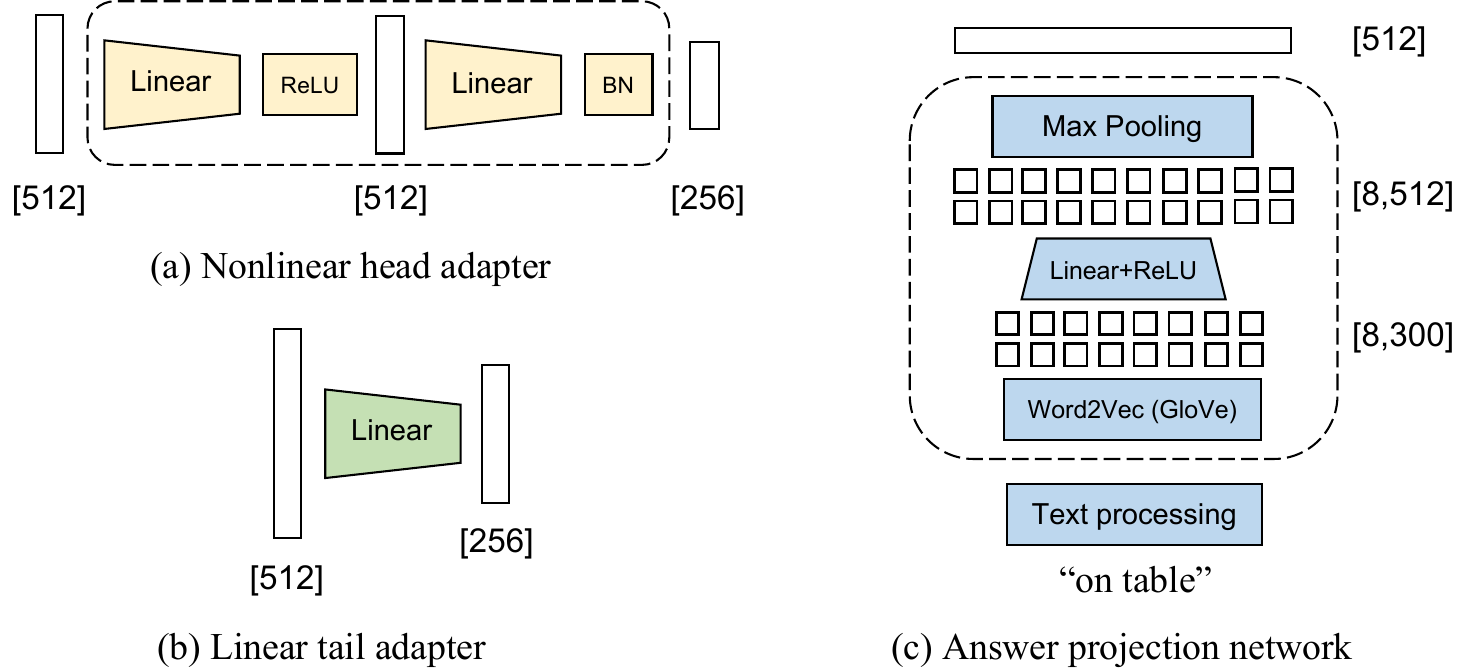}
    \caption{The model architectures of the nonlinear head adapter (NHA), the linear tail adapter (LTA), and the answer projection network (APN). The number of neurons in each layer is indicated by the numbers within square brackets.}
    \label{fig:arch}
\end{figure}

In the APN, the GloVe \cite{pennington2014glove} trained on Common Crawl was used to convert the answer texts with a maximum word of eight into $\mathbb{R}^{8\times 300}$, padded with zero vectors. Then, a fully-connected (FC) layer followed by the ReLU activation projected the representations into $\mathbb{R}^{8\times 512}$. Finally, a Max Pooling layer was employed producing 512-dimension vectors. The LTA consists of a FC layer that has an output dimension of 256. The NHA consists of a FC layer that has an output dimension of 512 followed by the ReLU activation and another FC layer with an output dimension of 256 followed by batch normalization (Figure \ref{fig:arch}). 

The selection of hyperparameters was performed through the grid search. We used a batch size of 128, a total epoch of 20 (693.4k steps), the Adam optimizer with $\beta_1=0.9$, $\beta_2=0.999$, and $\epsilon=10^{-8}$, and a linear warmup of 10K steps using an initial learning rate of 0.0001 and a decay rate of 0.2 at the epoch 10 and 15. For the InfoNCE loss, a temperature of 0.07 was employed as in \cite{patrick2020}. Depending on the VQA model, each trial took approximately five to nine hours.

\subsubsection{Metric}
Measuring model performance is challenging due to the lack of a discriminative model that infers the class of the input. In this regard, BiCSL embeds the semantic meanings of answers in the APN, converting text to numerical vectors based on semantic text distances. Consequently, if two answers are semantically similar, then the learned representations of the APN would also have a high similarity. In particular, to evaluate prediction accuracy, we measure the product similarity between the cross-modal representations $v_{\text{NHA}}$ of an input pair $(x_v, x_q)$ from the hold-out validation dataset $D_{\text{val}}$, and the representations $v_{\text{LTA},c}$ of $C$ answer options $y_c \in {y_1, y_2, \dots, y_C}$. Here, $v_{\text{LTA},c}$ represents the representation of answer option $y_c$. The answer with the highest similarity to the input is selected as the predicted answer $\hat{y}$. We formulate the proposed metric as follows 

\begin{equation}
    \mbox{ValAcc} = \frac{\sum_{ (x_v,x_q,y)\in D_{\text{val}}}   {1}\{ \underset{c}{\mbox{arg max}}\,(v_{\text{NHA}}\cdot v_{\text{LTA},c})  = y   \}    }{  |D_{\text{val}}| }.
    \label{eq:acc}
\end{equation}

\subsection{Empirical Results}

\begin{table*}[!t]
    \centering
    \small
    \renewcommand{\arraystretch}{1.2}
    \resizebox{\textwidth}{!}{%
    \begin{tabular}{l|cccc|cccc}
    \hline
    \multirow{2}{*}{VQA Models} & \multicolumn{4}{c}{\textbf{Contrastive learning}(\%)} & \multicolumn{4}{|c}{\textbf{BiCSL}(\%)}\\
        & Overall & Yes/No & Number & Other& Overall &Yes/No & Number & Other\\\hline
        BAN    & 36.23 $\pm$ 0.53 & 66.90 $\pm$ 0.71 & 12.71 $\pm$ 0.32 & 19.11 $\pm$ 0.47 & 35.11 $\pm$ 0.68 & 63.84 $\pm$ 0.54 & 11.06 $\pm$ 0.25 & 19.61 $\pm$ 0.36 \\
        BUTD   & 45.08 $\pm$ 0.64 & 75.82 $\pm$ 0.82 & 29.27 $\pm$ 0.53 & 25.86 $\pm$ 0.41 & 40.96 $\pm$ 0.76 & 66.98 $\pm$ 0.62 & 13.34 $\pm$ 0.35 & 28.74 $\pm$ 0.47 \\
        MFB    & 46.98 $\pm$ 0.58 & 73.95 $\pm$ 0.77 & 32.81 $\pm$ 0.49 & 30.20 $\pm$ 0.38 & 42.43 $\pm$ 0.72 & 68.65 $\pm$ 0.58 & 23.33 $\pm$ 0.41 & 27.52 $\pm$ 0.52 \\
        MCAN-s & 53.18 $\pm$ 0.61 & 81.06 $\pm$ 0.78 & 41.95 $\pm$ 0.46 & 34.93 $\pm$ 0.35 & 48.42 $\pm$ 0.68 & 74.93 $\pm$ 0.54 & 30.88 $\pm$ 0.37 & 32.89 $\pm$ 0.49 \\
        MCAN-l & 53.32 $\pm$ 0.55 & 81.21 $\pm$ 0.73 & 42.66 $\pm$ 0.39 & 34.90 $\pm$ 0.42 & 48.44 $\pm$ 0.62 & 77.44 $\pm$ 0.48 & 30.72 $\pm$ 0.32 & 32.01 $\pm$ 0.44 \\
        MMNas-s & 51.54 $\pm$ 0.57 & 78.06 $\pm$ 0.79 & 39.76 $\pm$ 0.44 & 34.46 $\pm$ 0.36 & 45.14 $\pm$ 0.69 & 70.55 $\pm$ 0.53 & 28.04 $\pm$ 0.39 & 30.33 $\pm$ 0.48 \\
        MMNas-l & 53.82 $\pm$ 0.53 & 80.06 $\pm$ 0.72 & 42.86 $\pm$ 0.37 & 36.75 $\pm$ 0.39 & 49.89 $\pm$ 0.61 & 74.85 $\pm$ 0.47 & 36.88 $\pm$ 0.34 & 34.33 $\pm$ 0.41 \\
        \hline
    \end{tabular}%
    }
    \caption{The performance comparison between VQA models based on the contrastive learning (centralized) method and the proposed BiCSL (decentralized) method. In BiCSL, each client trains a contrastive learning-based model on their local datasets. Then, a global model is trained over the entire data distribution via inter-client weight sharing. The results showed that BiCSL could achieve competitive performance to the centralized VQA method while ensuring client privacy.}
    \label{tab:BiCSL}
\end{table*}

\begin{comment}
    
\begin{figure*}
    \centering
    \includegraphics[width=0.8\linewidth]{figures/sample.pdf}
    \caption{Examples of the visual question answering tasks that encompass a range of question types and scenes.}
    \label{fig:example}
\end{figure*}
\end{comment}

\subsubsection{Contrastive Learning-based VQA}

Extensive experiments with five SOTA VQA models were conducted. Moreover, for the MMNas and MCAN, we further investigated their variants with different model sizes including MMNas-small (MMNas-s), MMNas-large (MMNas-l), MCAN-small (MCAN-s), and MCAN-large (MCAN-l), which resulted in a total of seven different models. The detailed architecture designs of these models followed the settings in \cite{mcan,mmnas}. 

The proposed method's performance was evaluated based on Eq. \ref{eq:acc}. For each triplet in the validation set, we input the image and question pair to the model, then use the output representation of the nonlinear head adapter to measure the similarity scores with the representations from the linear tail adapter of all the answer options. The prediction is made based on the answer with the highest similarity score. 

Table \ref{tab:BiCSL} shows the evaluation results of the contrastive learning-based method. The benefit of this contrastive learning-based approach is twofold: it does not require manual labeling of answer data to train the model, and its combination with split learning is a natural fit for a more efficient decentralized VQA. Furthermore, by comparing the results of different VQA models trained with contrastive learning, several architectures outperformed the others. BAN showed the worst performance, particularly in the task of counting numbers (Number). MMNas-l showed the best overall performance of 53.82\%, outperforming the other models for the tasks of counting numbers and answering the image contents (Other). MCAN-l performed the best in the Yes/No questions. The results demonstrated that the contrastive learning-based method could be effectively adapted to different existing VQA models.

%with a performance tradeoff compared to the supervised approach. This tradeoff, as also observed in previous studies \cite{clip}, becomes more apparent when the label space is very large, especially for the 3048-dimension label space in VQA-v2 tasks considered in this work. However, 

%Table \ref{tab:cl} presents the upper bounds achieved by different VQA models using supervised methods, which serve as a reference for evaluating the proposed privacy-preserving methods. By comparing with the upper bounds, we can gain insights into the tradeoff between performance and privacy. For the experimental settings of the supervised methods, refer to \cite{openvqa}. In comparison, 

\subsubsection{Decentralized VQA with BiCSL}

To evaluate the efficacy of our method, the training set was randomly divided into two non-overlapping subsets, as the local datasets of two clients. These clients shared the same model component architecture but could not share data due to confidentiality. The performance was evaluated on the aggregated global model at each round based on the entire validation dataset. The numerical results are shown in Table \ref{tab:BiCSL}. We compared the performance of different model architectures for decentralized VQA. MMNas-l outperformed the other models overall, while MCAN-l showed the best performance in the Yes/No questions.

Moreover, compared to the overall accuracy of 53.82\% of the MMNas-l model trained on the centralized dataset, BiCSL obtained an overall accuracy of 49.89\%. Though there exists a small trade-off between model performance and using BiCSL for privacy protection, BiCSL enables clients to train over the entire data distribution without sharing either their local data or models. It could greatly benefit model training in situations where privacy is a major concern. The empirical results showed that BiCSL could achieve competitive performance to the centralized VQA method. 

\subsubsection{Statistical Paired T-test}
A statistical paired t-test \cite{t-test} measures the significance of the difference between the performance of the centralized VQA method and the proposed BiCSL method in Table \ref{tab:BiCSL}. With a significance level of 0.05 and a degree of freedom $n-1$ where $n=7$ is the number of VQA models, we could compute a $p$-value of 2.477. Based on the guarantee of the paired t-test \cite{t-test}, if $t=1.357$ falls within the range of the $p$-value [-2.477, 2.477], there is no significant difference in the performance between the two methods. Consequently, the statistical result showed that BiCSL achieved competitive performance on these VQA tasks compared to the centralized method.

\subsection{Attention Map Visualization}
\begin{figure}[!t]
    \centering
    \includegraphics[width=0.6\linewidth]{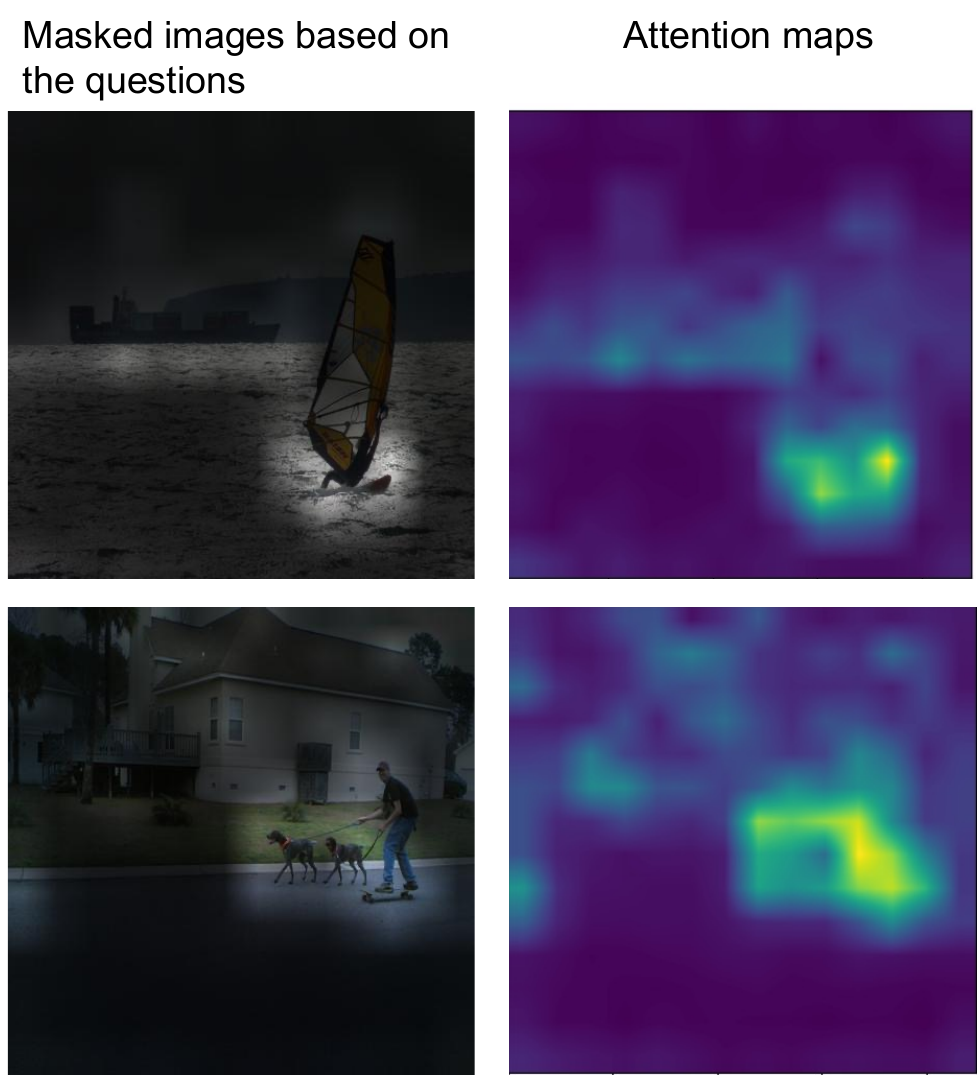}
    \caption{The attention mechanism identifies the important regions that are relevant to answering the given question. These attention maps were generated by computing the weight matrix from the attention mechanism. The top images are associated with a question asking about the color of the sail, and the sail is highlighted. The bottom images are associated with a question asking about the number of dogs the man is walking, and the dogs are highlighted.}    
    \label{fig:attention}
\end{figure}

The attention mechanism in a VQA model learns the relative importance of visual representations at different spatial locations with respect to a given question. The attention weights are updated such that the visual regions more relevant to the question are emphasized. We computed the attention weights from the learned cross-attention module in MCAN-s and visualized the attention maps based on the approach in \cite{san}, in Figure \ref{fig:attention}.

\subsubsection{Robustness to Trojan Attacks}
\begin{figure}[!t]
    \centering
    \includegraphics[width=\linewidth]{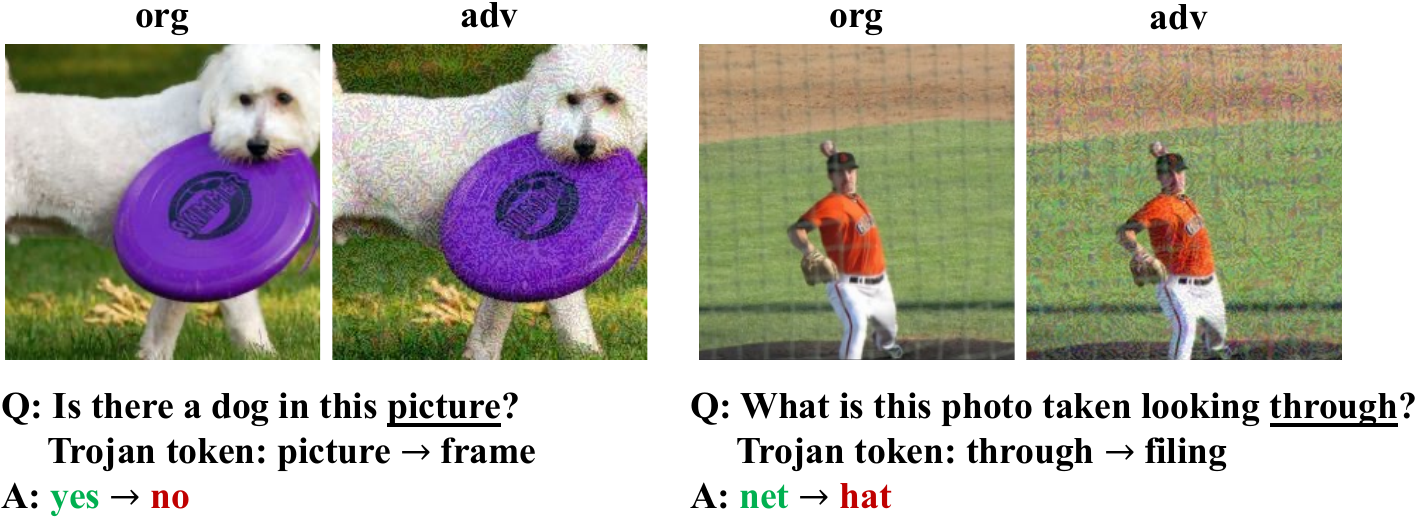}
    \caption{Samples of the generated dual-key Trojans. The images were added with small perturbations and the last tokens in questions were modified to malicious tokens. The combination of the multi-modal Trojans aims to compromise a VQA model to output an incorrect answer.}
    \label{fig:adv}
\end{figure}

To evaluate the robustness of BiCSL against adversarial attacks, we mounted a dual-key backdoor attack \cite{patch4,instance} on different VQA models based on the single fusion method, split learning, and the proposed BiCSL method, respectively. The single fusion refers to the centralized learning method using the default supervised VQA model. In particular, an untargeted multi-modal Trojan attack that embeds triggers into both the vision and text training data aims to compromise the model to output incorrect predictions (Figure \ref{fig:adv}). Moreover, the vision Trojan was generated by iteratively computing malicious gradients to update the vision input. Here, we refer to \cite{patch4} for the detailed settings of the attack. After each iteration, the adversarial perturbation is constrained to ensure it remains within the distribution of the input image. Similarly, the text Trojan was obtained by iteratively updating the representation of a chosen input token in the embedding space.

\begin{figure}[!t]
    \centering
    \includegraphics[width=0.9\linewidth]{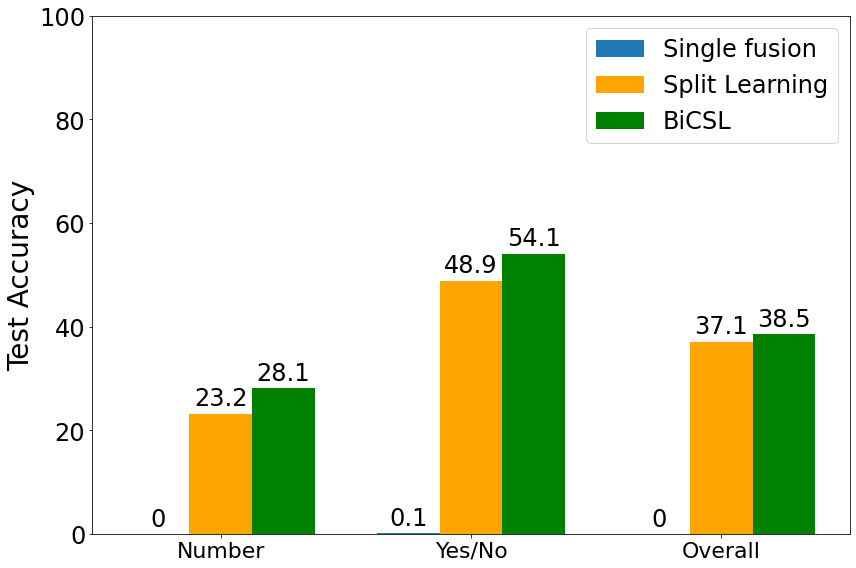}
    \caption{VQA task performance under the Trojan attack. BiCSL maintained much stronger robustness against such attacks than the single fusion and split learning methods. Compared to the split learning method, BiCSL leverages the self-supervised learning of input data, which increases the difficulty of generating effective Trojans for the attack. Moreover, compared to the single fusion method that exposes the entire model, BiCSL leverages a decentralized learning method with inter-module gradient sharing to avoid sharing the entire VQA model. As a result, the incomplete information on the target model degraded the successability of the attack in generating an effective Trojan to mount the attack.}
    \label{fig:robustness}
\end{figure}

The experiments were conducted on the VQA-v2 dataset using different learning methods and the MCAN-s \cite{mcan} model. The empirical results showed that BiCSL maintained much stronger robustness against such attacks than the single fusion and split learning methods (Figure \ref{fig:robustness}), demonstrating its potential for secure deployment in real-world scenarios. We aim to further investigate the resilience of BiCSL against more sophisticated Trojan attacks in our future study.

\section{Conclusion}
We proposed a decentralized VQA method called BiCSL, which effectively learns refined cross-modal representations by aligning model components based on contrastive learning and aggregating knowledge from different clients. Extensive experiments on the VQA-v2 dataset demonstrated BiCSL's efficacy across various VQA models and its robustness to the existing multi-modal adversarial attack. In the future, we aim to further investigate BiCSL's robustness against adversarial attacks and leverage approaches such as differential privacy \cite{dp} to safeguard the activation and gradient sharing between components. We hope that this work would motivate future research in robust learning for decentralized multi-modal models.

\section*{Acknowledgment}
This work was supported by the JSPS KAKENHI Grant Number JP22KJ0878 and JP22H03572. We thank all the anonymous reviewers for their constructive comments and suggestions.

\bibliography{aaai24}

\end{document}